\documentclass[10pt, conference]{IEEEtran} 

\usepackage{cite}
\usepackage{graphicx}
\usepackage{caption}
\graphicspath{ {./images/} }
\usepackage[thinlines]{easytable}
\usepackage{amsmath}
\usepackage{float}
\usepackage{threeparttablex}
\usepackage{algorithmic}
\usepackage{algorithm}
\usepackage{xspace}
\usepackage[T1]{fontenc}
\usepackage{subcaption}

\title{The More is not the Merrier: Investigating the Effect of Client Size on Federated Learning} 
\author{
\IEEEauthorblockN{Eleanor Wallach}
\IEEEauthorblockA{\textit{Department of Computer Science} \\
\textit{College of William \& Mary}\\
Williamsburg, VA \\
ewallach@wm.edu }
\and
\IEEEauthorblockN{Sage Siler}
\IEEEauthorblockA{
\textit{Department of Computer Science} \\
\textit{Appalachian State University}\\
Boone, NC \\
silerks@appstate.edu}
\and
\IEEEauthorblockN{Jing Deng}
\IEEEauthorblockA{
\textit{Department of Computer Science} \\
\textit{UNC Greensboro}\\
Greensboro, NC \\
jing.deng@uncg.edu }
}

\begin{document}

\maketitle

\begin{abstract}
Federated Learning (FL) has been introduced as a way to keep data local to clients while training a shared machine learning model, as clients train on their local data and send trained models to a central aggregator. It is expected that FL will have a huge implication on Mobile Edge Computing, the Internet of Things, and Cross-Silo FL. In this paper, we focus on the widely used FedAvg algorithm to explore the effect of the number of clients in FL. We find a significant deterioration of learning accuracy for FedAvg as the number of clients increases. To address this issue for a general application, we propose a method called Knowledgeable Client Insertion (KCI) that introduces a very small number of knowledgeable clients to the MEC setting. These knowledgeable clients are expected to have accumulated a large set of data samples to help with training. With the help of KCI, the learning accuracy of FL increases much faster even with a normal FedAvg aggregation technique. We expect this approach to be able to provide great privacy protection for clients against security attacks such as model inversion attacks. 
\end{abstract}


\section{Introduction}
\begin{figure}[h]
     \centering
     \includegraphics[width=3.5in]{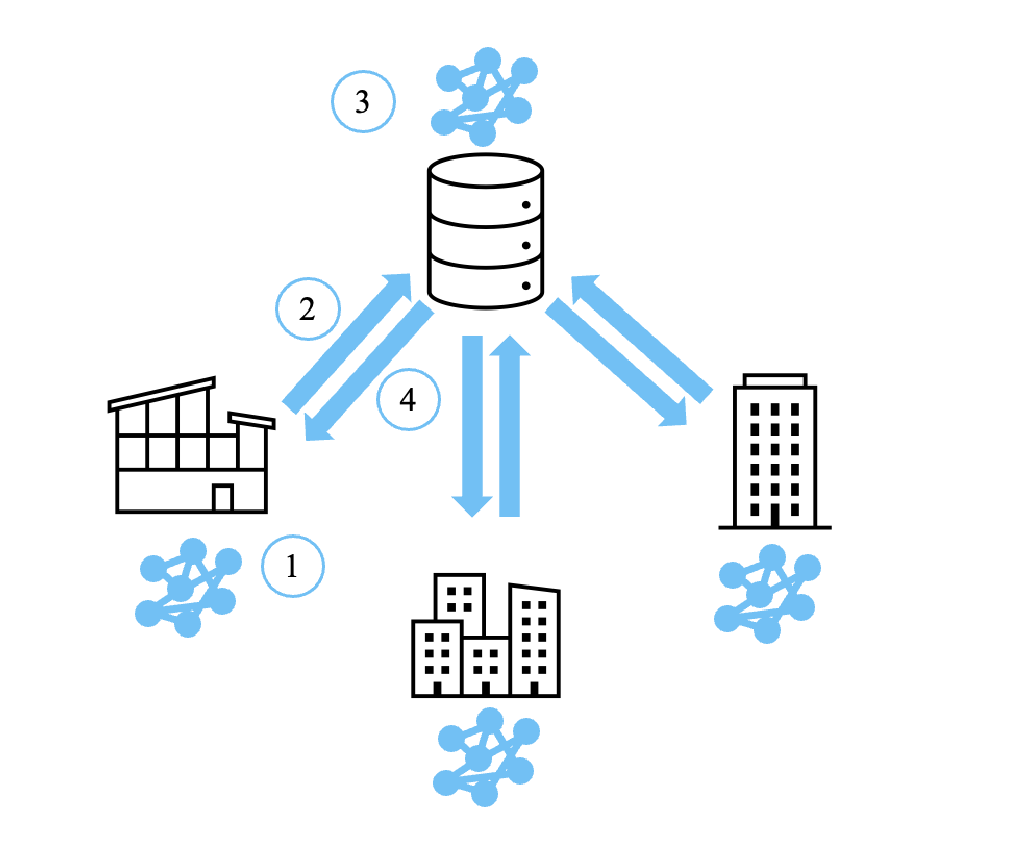}
     \caption{An Illustration of Federated Learning Framework.}
     \label{Figure:FL}
\end{figure}

Federated Learning (FL) was coined in 2017 by McMahan et al. as a different approach to machine learning~\cite{mcmahan17}. As outlined in Fig.~\ref{Figure:FL}, FL occurs when each client trains a machine learning model on their local data (1), sends the trained model's parameters to a central server (2), a central server aggregates trained models from multiple clients and creates a new global model (3) that is then sent to clients (4) where the process repeats~\cite{mcmahan17}. Among the many benefits of FL is its ability to protect local data. All training is done locally and none of the data used for training is sent to the global model, giving FL practical use in places such as banks and hospitals~\cite{lu24}. Privacy reasons like these are one reason why FedAvg was introduced, as it is one of the most prominent advantages of FL~\cite{mcmahan17}. Since the term's introduction, there has been intensive research on the topic. Two particular topics that have been looked into in FL are how the distribution of data affects the accuracy, and how the number of clients affects the accuracy. \par

Data can potentially be not identically and independently distributed (non-IID), which presents a problem for federated learning due to the local data of each client not always being representative of the global data. Thus, local updates are not necessarily in line with the global model, resulting in a lower test accuracy when using non-IID data compared to identically and independently distributed (IID) data \cite{li2022federated, Zhao18}. \par

Previous research has found conflicting information about the effect of the number of clients in FL. Li et al.~\cite{li2022federated} found the accuracy of all tested algorithms (FedAvg, FedProx, FedNova, and SCAFFOLD) decreased as the number of clients increased for multiple types of non-IID and IID data. Zhang et al.~\cite{zhang21} found the accuracy of FedAvg, FedProx, and their proposed CSFedAvg increased as the number of clients increased for non-IID data. Wong et al.~\cite{wong23} observed an increase in the accuracy of FedAvg as the number of clients increased. This improvement was more pronounced for non-IID data compared to IID data. Wu et al.~\cite{Wu21} found FedAvg to achieve a relatively stable accuracy across different numbers of clients with non-IID data. Li et al.~\cite{li2021model} considered 50 clients with a sample ratio of 1 and 100 client with a sample ratio of 0.2. They observed a decrease in accuracy with 100 clients compared to 50 clients for their proposed MOON algorithm and for FedAvg, FedProx, and SCAFFOLD. These papers suggest more research is needed into the effects of different amounts of clients in FL. \par 
In this paper, we explore how the number of clients impacts model accuracy. This rest of this paper is structured as follows. 
In Section~\ref{sec:related}, we review previous research relating to FL, client selection, and privacy. 
In Section~\ref{sec:background}, we discuss important applications of FL and briefly explain the FedAvg algorithm. 
In Section~\ref{sec:KCI}, we propose a new technique, KCI, to address the loss of accuracy with large numbers of clients. 
In Section~\ref{sec:results}, we outline our experimental conditions and evaluate our algorithm. We show the superiority of KCI even with large numbers of clients. 
In Section~\ref{sec:conclusion}, we summarize our new algorithm and provide future research directions.

\section{Related Work} \label{sec:related}
Many papers have worked off of FedAvg previously to propose new algorithms. 
Li et al.~\cite{li20} proposed FedProx and introduced a proximal term to better handle system and data heterogeneity. The proximal term controls the impact of local updates to the global model, achieving model updates more inline with the goals of the global model. Karimireddy et al.~\cite{Karimireddy19} proposed SCAFFOLD which uses cross client variance reduction to correct for "client-drift" by estimating the direction of the global model and each local model. Wang et al.~\cite{wang20} aggregated the normalized stochastic gradients from clients. Li et al.~\cite{li2021model} proposed MOON and used contrastive learning to improve FL. Clustering has also been proposed as a way to improve FL with non-IID data in particular~\cite{Li23Cluster, Ghosh20}. In clustering frameworks, clients are sorted into groups based on their data distribution, computation capabilities, and other factors. Model aggregation first occurs within each group or cluster. These intermediary model are aggregated by the server to create the global model. \par

Extensive research has looked into how to select clients to participate in each round of FL when data is non-IID and the sample ratio of clients is less than one. Zhang et al.~\cite{zhang21} proposed clients with larger degrees of IID data train the global model more frequently than clients with larger degrees of non-IID data. Wong et al.~\cite{wong23} considered the varying computational abilities of clients when selecting which clients would participate in training. Xin et al.~\cite{Xin22} proposed FCCPS which can change the number of clients chosen for training based on client performance. Nishio and Yonetani~\cite{Nishio19} proposed selecting clients for training based on their computational capacity and data distribution. \par

Data privacy in FL has been greatly researched as privacy is known to be a trade off of higher accuracies~\cite{chen23}. Privacy is built into the framework of FL since only model parameters are shared, yet there are still privacy concerns. Zhu et al.~\cite{Zhu19} showed that local, private data can be recreated from gradients. Since the gradients shared between clients and the server are public, there is risk of local data being exposed. This has led to some algorithms that aim to preserve clients' privacy while still ensuring high accuracy. Many papers utilize differential privacy (DP), a mathematical method to add noise to the system, as a way to increase privacy~\cite{chen23, zhao21, Triastcyn19}. DP allows for data statistics to be gathered about the dataset without revealing sensitive information about the data. \par

Client size has been investigated by some researchers in the past. For example, Zhang et al.~\cite{zhang21} explicitly point out an increase in accuracy as client size increases from 100 to 300 clients.
Song et al.~\cite{Song23} explicitly observe a decrease in accuracy as client size increases from 1 to 100 clients for IID data, or 10 to 100 clients for non-IID data. Xu et al.~\cite{xu23} also note that the accuracy drops as the number of clients increases from 100 to 500 clients. Li et al.~\cite{li2022federated} also confirm the drop of accuracy from 10 to 40 clients, and note that designing algorithms for a "large-scale setting with small data in the client" is a challenging problem that has yet to be solved. \par

Both Li et al.~\cite{li2022federated} and Xu et al.~\cite{xu23} test using multiple different algorithms, such as FedAvg, FedProx, and SCAFFOLD, along with their own algorithms, and the results are displayed in Table~\ref{Table:1}. It should be noted that all of these reports ran a different number of rounds, making the accuracies for each not completely comparable to the others, but regardless of the number of rounds for one particular test all ran until converging at the very least. These show that while there is a mixed consensus for accuracy decreasing, many reports show accuracy drops while increasing the number of clients. Xu et al.~\cite{xu23} discuss the importance of accuracy scaling well with client numbers since real-world applications tend to have a large amount of clients.

\begin{table}[t] 
\begin{center}
\caption{Test accuracy as number of clients increases}
\begin{tabular}{ |c|c c c c| } 

 \hline
& \multicolumn{4}{c|}{\textbf{Accuracy}} \\
   \textbf{\# of clients}        & Zhang \cite{zhang21} & Li \cite{li2022federated} & Song \cite{Song23} & Xu \cite{xu23}  \\
 \hline
 10 & - & ~0.68 & 0.503 & - \\
 20 & - & ~0.64 & - & - \\
 30 & - & ~0.61 & - & - \\
 40 & - & ~0.6 & - & - \\
 100 & ~0.575 & - & 0.414 & 0.5578 \\
 300 & ~0.645 & - & - & - \\
 500 & - & - & - & 0.3378 \\
 \hline
\end{tabular}
\label{Table:1} 
\end{center} 
\end{table} 

\section{Preliminary} \label{sec:background}
\subsection{Mobile Edge Computing}
Mobile Edge Computing (MEC) takes advantage of the computation ability of edge devices and edge servers in networks~\cite{Shi16edge}. Data is processed on edge devices or close to the device where it is gathered without the need to send data all the way to the cloud. This results in lower latency and higher bandwidth~\cite{lim2020edge}. To ensure data privacy, FL has been used in the context of MEC~\cite{lim2020edge}. In this setting, clients are edge devices and the central server is an edge device or edge server. Data never leaves edge devices where it is collected and used for model training. This also reduces communication latency and need for large upload bandwidth at the edge devices. 

\subsection{Cross-Silo Federated Learning}
FL can be divided into two main categories based on the classification of clients and the training data~\cite{huang22crosssilo}. In cross-device FL, clients are typically mobile devices or Internet of Things (IoT) devices. These devices generally have small local datasets to train with and varying computational abilities. The number of clients is extremely large (generally millions) and only a fraction of clients are chosen to participate in each round of training~\cite{kairouz21}. 
In cross-silo FL, clients are typically organizations or companies. The number of clients is smaller (generally 2-100) and the datasets are larger. The computational capacity of all clients is less diverse and there is full client participation in every round of training~\cite{huang22crosssilo}. Our work relates closely to cross-silo FL as we consider 100\% client participation across all rounds and a moderate number of clients with large datasets. 

\subsection{FedAvg Algorithm} 
The \texttt{FederatedAveraging} (FedAvg) algorithm was introduced in~\cite{mcmahan17} and has been used extensively. It is frequently used as a point of comparison when proposing new algorithms~\cite{li20, li2021model, zhang21, wang20}. 
In FedAvg, as outlined in Algorithm~\ref{alg:one}, each client computes the average gradient using its local data and sends this model to the server. The server then aggregates clients' gradients with weights. These weights $\frac{n_k}{m_t}$ represent the amount of data samples a client has in relation to the total number of data samples. This updates the global model, which is sent to clients and the process repeats for each communication round. We focus on FL using FedAvg in this work and leave the investigation toward other aggregation techniques to our future.  

\begin{algorithm}[t]
\caption{FedAvg. $T$ is the number of communication rounds, $K$ is the number of clients, $E$ is the number of local epochs, $\eta$ is the learning rate, $B$ is the mini-batch size, $l$ is the loss function} \label{alg:one}
\begin{algorithmic}[1]
\STATE \textbf{Server executes:}
\STATE initialize global model $w_0$
\FOR{each communication round $t = 1, 2,...,T$}
\STATE $\mathcal{S}_t \gets$ set of $K$ clients 
\FOR{each client $k \in \mathcal{S}_t$ \textbf{in parallel}}
\STATE $w_{t+1}^k \gets $ ClientTraining($k$, $w_t$)
\ENDFOR
\STATE $m_{t} \gets \sum_{k \in \mathcal{S}_t} n_k$
\STATE $w_{t+1} \gets \sum_{k \in \mathcal{S}_t} \frac{n_k}{m_t}w_{t+1}^k $
\ENDFOR
\STATE
\STATE \textbf{ClientTraining($k$, $w$):}
\STATE $\mathcal{B} \gets$ split local dataset into batches of size $B$
\FOR{each epoch $e = 1,2,...,E$}
\FOR{each batch $b \in \mathcal{B}$ } 
\STATE $w \gets w - \eta \nabla l (w;b)$ 
\ENDFOR
\ENDFOR
\STATE \textbf{Return} $w$ 
\end{algorithmic}
\end{algorithm}

\subsection{Accuracy drop due to increased number of clients} 

\begin{figure}[t]
     \centering
     \includegraphics[width=1\linewidth]{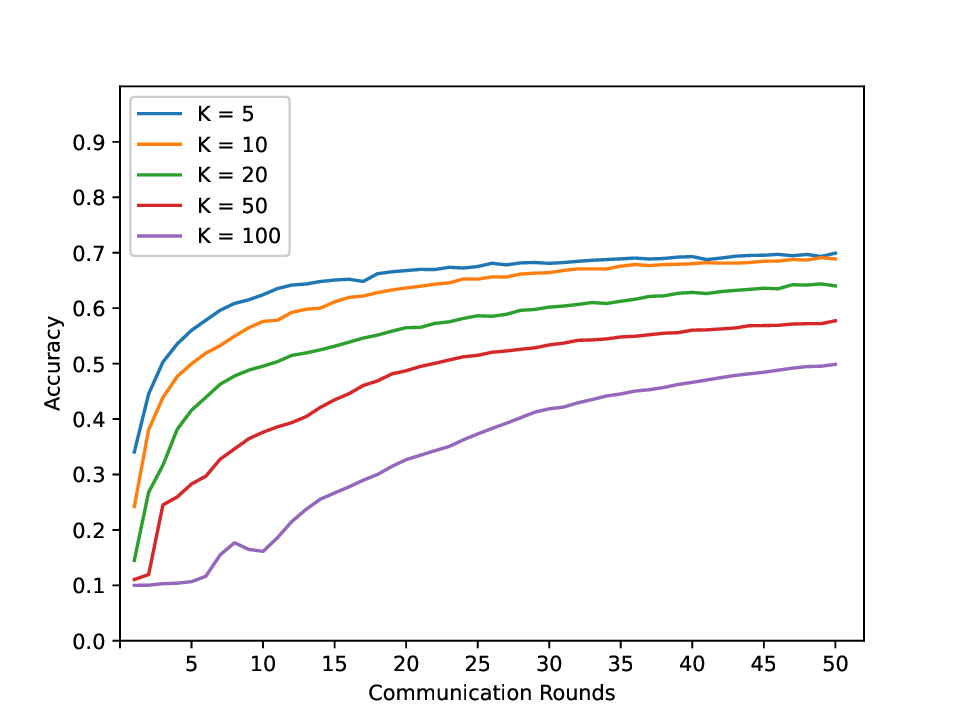}
     \caption{Accuracy for different $K$ of FedAvg}
     \label{Figure:prelim}
\end{figure}

Our preliminary study focuses on FL with different numbers of clients in the learning process. We differ the number of clients $K \in \{5,10,20,50,100\}$. Fig.~\ref{Figure:prelim} shows the change in accuracy as the number of clients increases. We observe a significant decrease in test accuracy as $K$ increases from 5 to 100. The final accuracy for 100 clients compared to 5 clients suggests an approximate 30\% decrease in test accuracy. This leads us to explore how to maintain a consistent accuracy as the number of clients increases.

\section{Knowledgeable Client Insertion (KCI)} \label{sec:KCI}

In this work, we propose a technique called Knowledgeable Client Insertion (KCI) to intrinsically improve the learning accuracy of FL. The intuitive explanation of the design is based on our observation that most clients in FL are distributed at different locations with various and mostly limited data for the entire learning procedure. An artificial insertion of a few clients, sometimes even one, that carry most of the potential training data, both in size and categories, would significantly improve the rate of learning. This will also allow privacy-concerned clients to maintain the secrecy of their data samples. A more thorough investigation would be needed for privacy leakage toward passive and active attackers~\cite{Xiong21}, but we leave it to our future work.\par
In KCI, $m$ new clients are inserted into the FL framework and they are distributed throughout the entire system. Each of these clients has a local dataset proportional to the overall size of the dataset. We use $\lambda$ to regulate the amount of data these clients have, with $\lambda = 1$ representing the client having all local data samples and $\lambda = 0$ representing the client having no local data. The data samples the inserted client(s) receive can be obtained through outside channels. 

In practice, we expect such knowledgeable clients to be physically deployed and inserted into the MCE domains as the FL procedure goes underway. Furthermore, it is possible to maneuver existing clients and turn them into knowledgeable clients by injecting more data samples, a process that warrants further investigation.

\section{Experimental Results} \label{sec:results}

\subsection{Setup}

We run simulations using code for FL using FedAvg~\cite{li2022federated} and modify it to implement KCI. We use the CIFAR-10 dataset~\cite{cifar10}, as it has been shown to yield lower accuracy than MNIST, suggesting it is more complex~\cite{zhang21, li2021model, wong23, mcmahan17, Nishio19}. CIFAR-10 contains 60,000 32x32 images spread evenly across 10 classes; 50,000 images are used as training data and the remaining 10,000 are used as testing data. We partition data uniformly amongst clients as IID. \par
We use a Convolutional Neural Network (CNN) with the same architecture as~\cite{li2022federated}. The CNN has two 5x5 convolution layers (6 channels for the first and 16 for the second) each followed by a 2x2 max pooling layer and two fully connected layers (the first with 120 units and ReLU activation and the second with 84 units and ReLU activation). \par
Unless otherwise stated, the mini-batch size $B$ is 64, the learning rate $\eta$ is 0.01, momentum is 0.9, the number of local epochs $E$ is 5, the number of communication rounds $T$ is 50, and the sample ratio of clients is 1. We differ the number of clients $K\in\{10,20,100\}$.%
\footnote{While we acknowledge that these experimental settings are rather limited, we expect our results and conclusions on hold on other large datasets, e.g., CIFAR-10 or CIFAR-100, and on comparisons with other FL aggregation techniques. We leave the evaluations of those to our future work.}

\subsection{Evaluation of KCI}

\begin{figure}[t]
     \centering
     \includegraphics[width=1\linewidth]{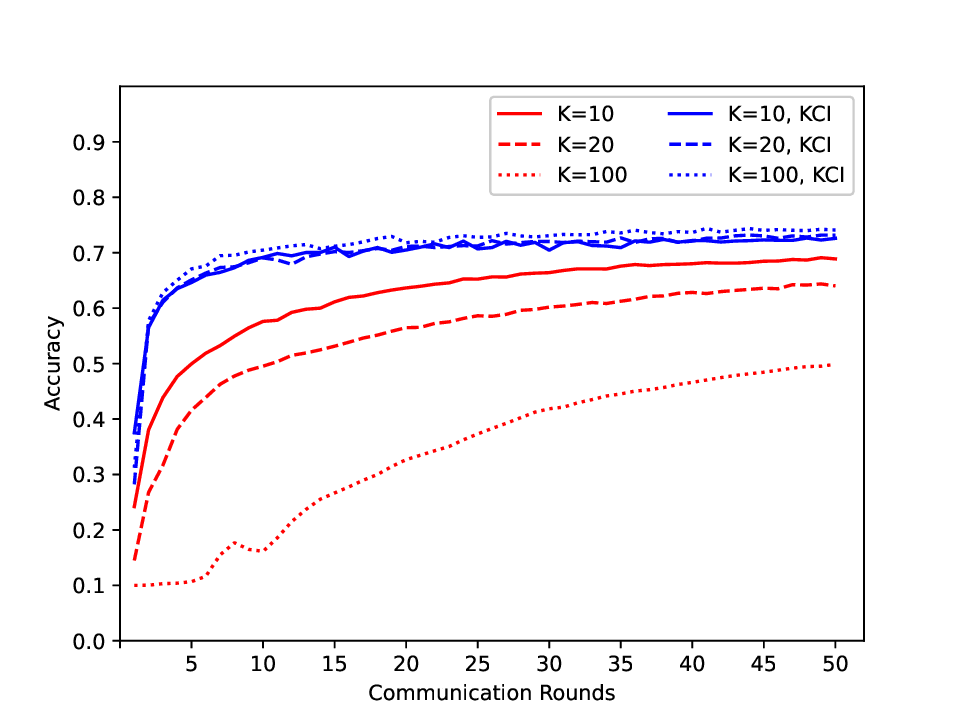}
     \caption{Accuracy comparison between FedAvg and KCI differing $K$ ($m=1$ and $\lambda=1$).}
     \label{Figure:KCI_FedAvg}
\end{figure}

We run simulations investigating the accuracy of KCI and compare it with FedAvg. Fig.~\ref{Figure:KCI_FedAvg} shows the accuracy when we insert $m = 1$ clients with $\lambda = 1$. The final accuracy of KCI is very consistent across different values of $K$. Allowing one client to train with all of the data greatly improves the accuracy for 100 clients. This is due to the nature in which data is partitioned amongst clients. When the dataset is split amongst 100 clients, each client receives much less data (and potentially even lower number of categories) to train compared to when the dataset is split amongst 10 or 20 clients. Overall, KCI yields a 45\% accuracy increase for 100 clients, a 16\% increase with 20 clients and a 5\% increase for 10 clients. 

\begin{figure}[t]
    \centering
    \includegraphics[width=3.5in]{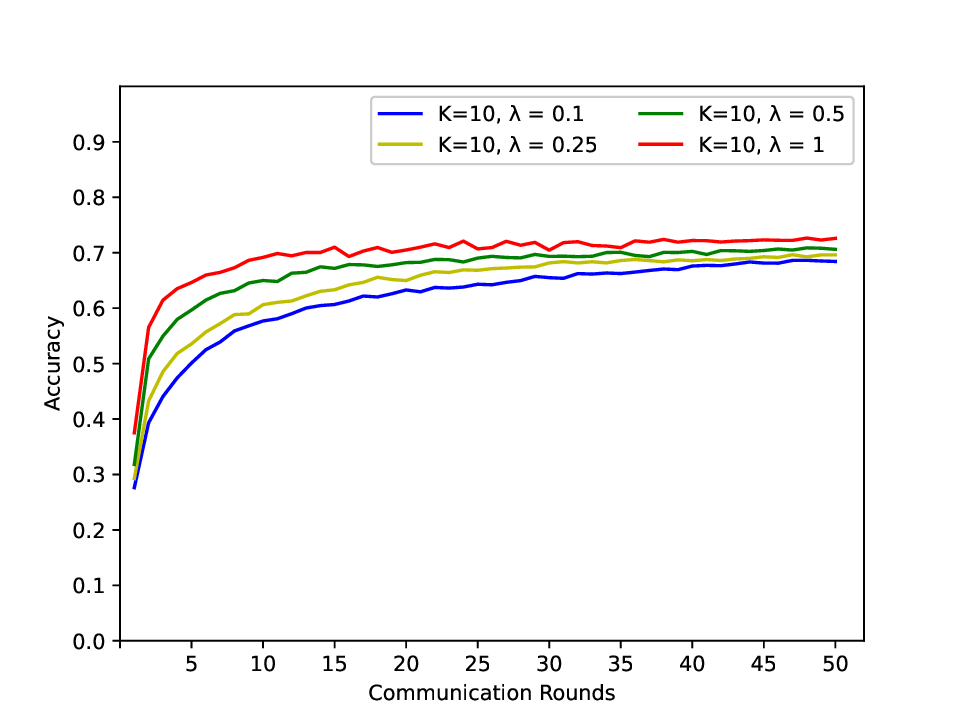}
    \caption{Accuracy for different $\lambda$ for KCI ($m=1$ and $K=10$).}
    \label{Figure:lambda}
\end{figure}

We then change the proportion of the training dataset that the artificial client receives. 
Fig.~\ref{Figure:lambda} shows the affect of different values of $\lambda$ on KCI. With $\lambda = 1$, the artificial client receives the entire training dataset for training. When $\lambda = 0.5,0.25,0.1$, the client receives half, a quarter, and a tenth of the training dataset, respectively. Having more data local to the artificial client leads to higher accuracy. 

\begin{figure}[t]
    \centering
    \includegraphics[width=3.5in]{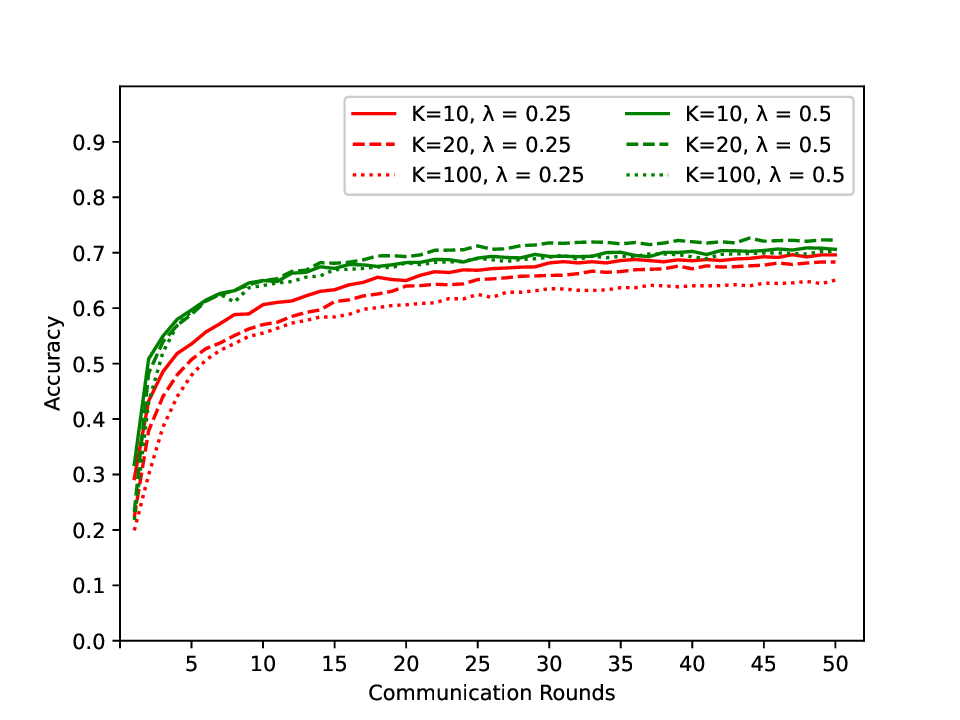}
    \caption{Accuracy for different $\lambda$ and $K$ ($m=1$).}
    \label{Figure:lambda-diffK}
\end{figure}

The change in accuracy does not appear to be proportional to the value of $\lambda$. We show accuracy values for different $K$ and $\lambda$ in Fig.~\ref{Figure:lambda-diffK}. For $K=100$, $\lambda = 1$ yields a final accuracy of $0.732$ while $\lambda = 0.5$ yields a final accuracy of $0.702$. This suggests that even if the artificial client does not have access to the entire dataset and only has access to half or part of the dataset, the accuracy improvement is still significant. 

\begin{figure}[h]
    \centering
    \includegraphics[width=3.5in]{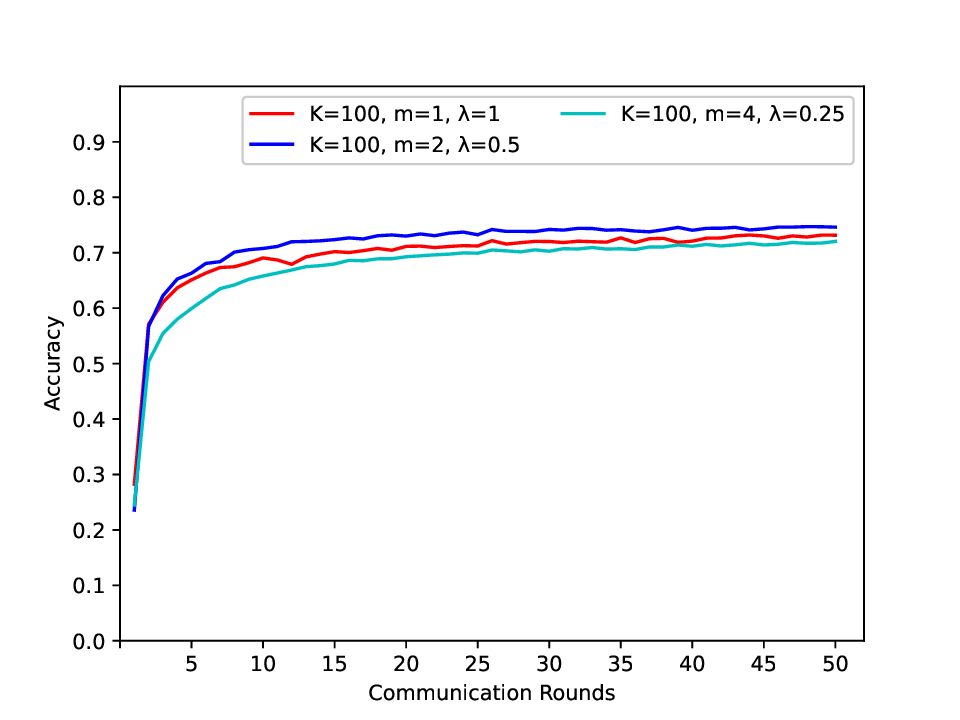}
    \caption{Accuracy for different $m$ and $\lambda$ ($K=100$).}
    \label{Figure:diff-m-1}
\end{figure}

\begin{figure}[h]
    \centering
    \includegraphics[width=3.5in]{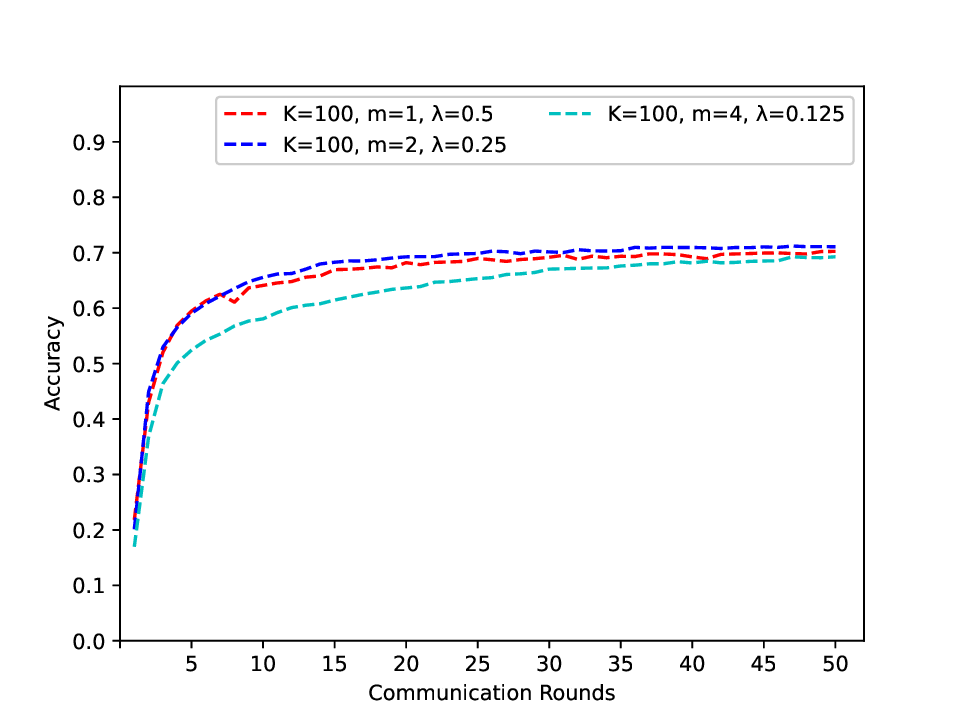}
    \caption{Accuracy for different $m$ and $\lambda$ ($K=100$).}
    \label{Figure:diff-m-0.5}
\end{figure}

We consider how data is dispersed amongst artificial clients. Fig.~\ref{Figure:diff-m-1} shows the the affect of $m$ and $\lambda$ when all local data is available to be used by artificial client(s). When one client receives all of the local data ($m = 1$, $\lambda = 1$), the accuracy is lower compared to when this data is evenly divided amongst two clients ($m = 2$, $\lambda = 0.5$). Dividing the data amongst four clients ($m = 4$) with $\lambda = 0.25$ yields a lower accuracy compared to using one or two clients. This trend is further supported when considering if artificial client(s) only have access to half of the local dataset. Fig.~\ref{Figure:diff-m-0.5} shows different values of $m$ and $\lambda$ when only half of the local training data is available to the artificial client(s). With artificial clients only using half of the local dataset, using $m = 2$ results in the best accuracies. This suggests setting $m = 2$ regardless of the amount of local data available to be split amongst clients yields the highest accuracies.

\section{Conclusion \& Future Directions} \label{sec:conclusion}
The FL algorithm, FedAvg, experiences a lower accuracy as the number of clients increases. This is due to the nature in which data is dispersed amongst clients. To combat this, we propose KCI which utilizes artificial client insertion. We find KCI to achieve up to 40\% higher accuracy compared to FedAvg for scenarios with large numbers of clients.

Future research will include an in depth privacy protection, KCI's interactions with other FL aggregation techniques, and KCI's performance under various non-IID data settings. 

\section{Acknowledgment}
This research was partially supported by NSF grant 2349369.

\bibliographystyle{IEEEtran}
\bibliography{finalreport}

\begin{thebibliography}{10}
\providecommand{\url}[1]{#1}
\csname url@samestyle\endcsname
\providecommand{\newblock}{\relax}
\providecommand{\bibinfo}[2]{#2}
\providecommand{\BIBentrySTDinterwordspacing}{\spaceskip=0pt\relax}
\providecommand{\BIBentryALTinterwordstretchfactor}{4}
\providecommand{\BIBentryALTinterwordspacing}{\spaceskip=\fontdimen2\font plus
\BIBentryALTinterwordstretchfactor\fontdimen3\font minus \fontdimen4\font\relax}
\providecommand{\BIBforeignlanguage}[2]{{%
\expandafter\ifx\csname l@#1\endcsname\relax
\typeout{** WARNING: IEEEtran.bst: No hyphenation pattern has been}%
\typeout{** loaded for the language `#1'. Using the pattern for}%
\typeout{** the default language instead.}%
\else
\language=\csname l@#1\endcsname
\fi
#2}}
\providecommand{\BIBdecl}{\relax}
\BIBdecl

\bibitem{mcmahan17}
\BIBentryALTinterwordspacing
B.~McMahan, E.~Moore, D.~Ramage, S.~Hampson, and B.~A.~y. Arcas, ``{Communication-Efficient Learning of Deep Networks from Decentralized Data},'' in \emph{Proceedings of the 20th International Conference on Artificial Intelligence and Statistics}, ser. Proceedings of Machine Learning Research, A.~Singh and J.~Zhu, Eds., vol.~54.\hskip 1em plus 0.5em minus 0.4em\relax PMLR, 20--22 Apr 2017, pp. 1273--1282. [Online]. Available: \url{https://proceedings.mlr.press/v54/mcmahan17a.html}
\BIBentrySTDinterwordspacing

\bibitem{lu24}
Z.~Lu, H.~Pan, and Y.~Dai, ``Federated learning with non-iid data: A survey,'' \emph{IEEE INTERNET OF THINGS JOURNAL}, 2024.

\bibitem{li2022federated}
Q.~Li, Y.~Diao, Q.~Chen, and B.~He, ``Federated learning on non-iid data silos: An experimental study,'' in \emph{IEEE International Conference on Data Engineering}, 2022.

\bibitem{Zhao18}
Y.~Zhao, M.~Li, L.~Lai, N.~Suda, D.~Civin, and V.~Chandra, ``Federated learning with non-iid data,'' 2018.

\bibitem{zhang21}
W.~Zhang, X.~Wang, P.~Zhou, W.~Wu, and X.~Zhang, ``Client selection for federated learning with non-iid data in mobile edge computing,'' in \emph{IEEE Access}, 2021.

\bibitem{wong23}
\BIBentryALTinterwordspacing
K.-S. Wong, M.~Nguyen-Duc, K.~Le-Huy, L.~Ho-Tuan, C.~Do-Danh, and D.~Le-Phuoc, ``An empirical study of federated learning on iot-edge devices: Resource allocation and heterogeneity,'' 2023. [Online]. Available: \url{https://arxiv.org/abs/2305.19831}
\BIBentrySTDinterwordspacing

\bibitem{Wu21}
P.~Wu, T.~Imbiriba, J.~Park, S.~Kim, and P.~Closas, ``Personalized federated learning over non-iid data for indoor localization,'' in \emph{2021 IEEE 22nd International Workshop on Signal Processing Advances in Wireless Communications (SPAWC)}, 2021, pp. 421--425.

\bibitem{li2021model}
Q.~Li, B.~He, and D.~Song, ``Model-contrastive federated learning,'' in \emph{Proceedings of the IEEE/CVF Conference on Computer Vision and Pattern Recognition}, 2021.

\bibitem{li20}
T.~Li, A.~K. Sahu, M.~Zaheer, M.~Sanjabi, A.~Talwalkar, and V.~Smith, ``Federated optimization in heterogeneous networks,'' 2020.

\bibitem{Karimireddy19}
\BIBentryALTinterwordspacing
S.~P. Karimireddy, S.~Kale, M.~Mohri, S.~J. Reddi, S.~U. Stich, and A.~T. Suresh, ``{SCAFFOLD:} stochastic controlled averaging for on-device federated learning,'' \emph{CoRR}, vol. abs/1910.06378, 2019. [Online]. Available: \url{http://arxiv.org/abs/1910.06378}
\BIBentrySTDinterwordspacing

\bibitem{wang20}
J.~Wang, Q.~Liu, H.~Liang, G.~Joshi, and H.~V. Poor, ``Tackling the objective inconsistency problem in heterogeneous federated optimization,'' 2020.

\bibitem{Li23Cluster}
X.~Li, Y.~Zhao, and C.~Qiao, ``Rcsr: Robust client selection and replacement in federated learning,'' in \emph{2023 IEEE 29th International Conference on Parallel and Distributed Systems (ICPADS)}, 2023, pp. 1577--1584.

\bibitem{Ghosh20}
A.~Ghosh, J.~Chung, D.~Yin, and K.~Ramchandran, ``An efficient framework for clustered federated learning,'' in \emph{Advances in Neural Information Processing Systems}, H.~Larochelle, M.~Ranzato, R.~Hadsell, M.~Balcan, and H.~Lin, Eds., vol.~33.\hskip 1em plus 0.5em minus 0.4em\relax Curran Associates, Inc., 2020, pp. 19\,586--19\,597.

\bibitem{Xin22}
F.~Xin, J.~Zhang, J.~Luo, and F.~Dong, ``Federated learning client selection mechanism under system and data heterogeneity,'' in \emph{2022 IEEE 25th International Conference on Computer Supported Cooperative Work in Design (CSCWD)}, 2022, pp. 1239--1244.

\bibitem{Nishio19}
\BIBentryALTinterwordspacing
T.~Nishio and R.~Yonetani, ``Client selection for federated learning with heterogeneous resources in mobile edge,'' in \emph{ICC 2019 - 2019 IEEE International Conference on Communications (ICC)}.\hskip 1em plus 0.5em minus 0.4em\relax IEEE, May 2019. [Online]. Available: \url{http://dx.doi.org/10.1109/ICC.2019.8761315}
\BIBentrySTDinterwordspacing

\bibitem{chen23}
\BIBentryALTinterwordspacing
H.~Chen, T.~Zhu, T.~Zhang, W.~Zhou, and P.~S. Yu, ``Privacy and fairness in federated learning: On the perspective of tradeoff,'' \emph{ACM Comput. Surv.}, vol.~56, no.~2, sep 2023. [Online]. Available: \url{https://doi.org/10.1145/3606017}
\BIBentrySTDinterwordspacing

\bibitem{Zhu19}
L.~Zhu, Z.~Liu, and S.~Han, ``Deep leakage from gradients,'' in \emph{Advances in Neural Information Processing Systems}, H.~Wallach, H.~Larochelle, A.~Beygelzimer, F.~d\textquotesingle Alch\'{e}-Buc, E.~Fox, and R.~Garnett, Eds., vol.~32.\hskip 1em plus 0.5em minus 0.4em\relax Curran Associates, Inc., 2019.

\bibitem{zhao21}
Y.~Zhao, J.~Zhao, M.~Yang, T.~Wang, N.~Wang, L.~Lyu, D.~Niyato, and K.-Y. Lam, ``Local differential privacy-based federated learning for internet of things,'' \emph{IEEE Internet of Things Journal}, vol.~8, no.~11, pp. 8836--8853, 2021.

\bibitem{Triastcyn19}
\BIBentryALTinterwordspacing
A.~Triastcyn and B.~Faltings, ``Federated learning with bayesian differential privacy,'' in \emph{2019 IEEE International Conference on Big Data (Big Data)}.\hskip 1em plus 0.5em minus 0.4em\relax IEEE, Dec. 2019. [Online]. Available: \url{http://dx.doi.org/10.1109/BigData47090.2019.9005465}
\BIBentrySTDinterwordspacing

\bibitem{Song23}
R.~Song, D.~Liu, D.~Z. Chen, A.~Festag, C.~Trinitis, M.~Schulz, and A.~Knoll, ``Federated learning via decentralized dataset distillation in resource-constrained edge environments,'' 2023.

\bibitem{xu23}
J.~Xu, M.~Yang, W.~Ding, and S.-L. Huang, ``Stabilizing and improving federated learning with non-iid data and client dropout,'' 2023.

\bibitem{Shi16edge}
W.~Shi, J.~Cao, Q.~Zhang, Y.~Li, and L.~Xu, ``Edge computing: Vision and challenges,'' \emph{IEEE Internet of Things Journal}, vol.~3, no.~5, pp. 637--646, 2016.

\bibitem{lim2020edge}
W.~Y.~B. Lim, N.~C. Luong, D.~T. Hoang, Y.~Jiao, Y.-C. Liang, Q.~Yang, D.~Niyato, and C.~Miao, ``Federated learning in mobile edge networks: A comprehensive survey,'' 2020.

\bibitem{huang22crosssilo}
C.~Huang, J.~Huang, and X.~Liu, ``Cross-silo federated learning: Challenges and opportunities,'' 2022.

\bibitem{kairouz21}
{P. Kairouz {\it et al.}}, ``Advances and open problems in federated learning,'' 2021, arXiv: 1912.04977.

\bibitem{Xiong21}
Z.~Xiong, Z.~Cai, D.~Takabi, and W.~Li, ``Privacy threat and defense for federated learning with non-iid data in aiot,'' \emph{IEEE Transactions on Industrial Informatics}, vol.~18, no.~2, pp. 1310--1321, 2021.

\bibitem{cifar10}
A.~Krizhevsky, ``Learning multiple layers of features from tiny images,'' 2009.

\end{thebibliography}

\end{document}